# Machine learning-based algorithms for at-home respiratory disease monitoring and respiratory assessment

Negar Orangi-Fard, PhD[1], Alexandru Bogdan PhD[2], Hersh Sagreiya, MD[3]

[1]Georgia Gwinnett College, Lawrenceville, GA; [2]CereVu Medical, San Francisco, CA; [3]University of Pennsylvania, Philadelphia, PA

*Abstract—* Respiratory diseases impose a significant burden on global health, with current diagnostic and management practices primarily reliant on specialist clinical testing. This work aims to develop machine learning-based algorithms to facilitate at-home respiratory disease monitoring and assessment for patients undergoing continuous positive airway pressure (CPAP) therapy. Data were collected from 30 healthy adults, encompassing respiratory pressure, flow, and dynamic thoraco-abdominal circumferential measurements under three breathing conditions: normal, panting, and deep breathing. Various machine learning models, including the random forest classifier, logistic regression, and support vector machine (SVM), were trained to predict breathing types. The random forest classifier demonstrated the highest accuracy, particularly when incorporating breathing rate as a feature. These findings support the potential of AI-driven respiratory monitoring systems to transition respiratory assessments from clinical settings to home environments, enhancing accessibility and patient autonomy. Future work involves validating these models with larger, more diverse populations and exploring additional machine learning techniques.

*Clinical Relevance—* Early and accurate detection of respiratory conditions during CPAP therapy can improve patient outcomes and reduce burden on the healthcare system. This work supports the potential of AI-driven respiratory monitoring to enhance the management of respiratory diseases at home, allowing for timely intervention and better long-term care.

## 1. INTRODUCTION

Respiratory diseases, including asthma, chronic obstructive pulmonary disease (COPD), and sleep apnea, are among the leading causes of morbidity and mortality worldwide, posing a substantial burden on healthcare systems and affected individuals [1, 2]. Traditional diagnostic and management strategies for these conditions depend heavily on specialist clinical testing conducted in controlled environments. Such reliance on clinical settings limits accessibility, delays diagnosis, and can hinder continuous monitoring, which is crucial for effective disease management [3].

Recent advancements in sensing technology and machine learning offer promising avenues to shift respiratory testing from specialized clinical environments to home and primary care settings. This transition can enhance access to care, provide real-time monitoring, and empower patients with actionable health insights. As an example, Karlen et al. (2010) demonstrated that machine learning models could effectively monitor sleep apnea at home using wearable sensors [4]. Similarly, Alvarez-Estevez and Moret-Bonillo (2015) achieved significant success in classifying respiratory patterns associated with sleep-disordered breathing using machine learning techniques [5]. Such advancements highlight the potential of these technologies to improve respiratory health management by enabling continuous monitoring.

Continuous positive airway pressure (CPAP) therapy, commonly used in the treatment of obstructive sleep apnea, presents an ideal context for integrating machine learning algorithms to monitor and assess respiratory health remotely.

The primary objective of this work is to develop and evaluate machine learning-based algorithms capable of predicting breathing types—normal, panting, and deep—in individuals using CPAP therapy. By leveraging non-invasive measurements such as respiratory pressure, flow, and thoraco-abdominal circumference, the proposed system aims to provide accurate and reliable respiratory assessments outside of clinical settings.

## 2. METHODS

In this section, the proposed methodology for remotely monitoring respiratory disease, as well as the assessment for patients undergoing CPAP therapy using machine learning techniques, is described.

### A. Data

The following dataset was used in this work:

"Pressure, Flow, and Dynamic Thoraco-Abdominal Circumference Data for Adults Breathing Under CPAP Therapy," a dataset publicly available on PhysioNet [1]:

A comprehensive dataset was collected from 30 healthy adult participants at the University of Canterbury, adhering to the ethical standards set by the Human Research Ethics Committee (HREC). The demographic distribution included an equal split of 15 males and 15 females, aged between 19 and 37 years. Participants were recruited via advertisements and provided informed consent prior to participation.

Breathing Protocols [1]:

Participants were instructed to perform three distinct breathing patterns: normal breathing, panting as rapidly as possible, and deep breathing. Each breathing trial was conducted using a CPAP mask (FreeMotion RT041, Fisher and Paykel Healthcare, New Zealand) connected to a HEPA filter (HMEF 002874, Smiths Medical, Minneapolis, MN, USA). The custom bi-directional Venturi device was integrated into the breathing circuit to measure respiratory pressure and flow.

   Normal Breathing: 65-second duration

   Panting: 35-second duration

   Deep Breathing: 65-second duration

   Participants determined their rest intervals between trials to ensure comfort and reduce fatigue.

Measurement Devices [1]:

Dynamic thoracic and abdominal circumferential monitoring bands were placed around the participants' thorax and abdomen at the level of the armpit and waist, respectively. A MATLAB script (MATLAB 2020a, The MathWorks Inc, Natick, MA, USA) was utilized to sample data at a frequency of 100Hz for each trial segment, capturing variables such as breathing rate and positive end-expiratory pressure (PEEP) settings.

Demographic Data [1]:

A comprehensive questionnaire was administered to collect demographic information, including sex, height, weight, age, smoking or vaping history, and asthmatic status. Participants also reported their frequency of using medications for asthma and any perceived changes in breathing difficulty during the trials.

### B. *Data Preprocessing and Feature Extraction*

The collected dataset included measurements of pressure at the Venturi throat ($cmH_2O$), flow (L/s), tidal volume (L), and the circumferences of the chest and abdomen (mm). Initial preprocessing steps involved:

- *Data Cleaning:* Removal of rows containing NaN values to ensure data integrity.
- *Breathing Rate Estimation:* An FFT (Fast Fourier Transform) algorithm was employed to estimate the breathing rate from the respiratory measurements. This derived feature was incorporated into the machine learning models to enhance predictive performance.

- *Normalization:* Feature scaling was applied to standardize the range of the input variables, facilitating more effective model training.

C. **Machine Learning Methods:**

The performance of three different machine learning techniques was compared in their ability to remotely monitor respiratory disease and assess breathing patterns in patients undergoing CPAP therapy:

- *Support vector machine (SVM)*:

The Support Vector Machine (SVM) machine learning model transforms the original feature space into a higher-dimensional space where it identifies the optimal hyperplane that maximizes the margin between the closest data points from different classes. In this work, a Gaussian radial basis function (RBF) kernel was employed to enable the SVM to handle non-linear relationships within the data [4].

- *Random Forest Classifier:*

This is an ensemble learning method that constructs multiple decision trees during training and outputs the mode of the classes for classification tasks. Parameters such as the number of trees and maximum depth were optimized using cross-validation techniques [5].

- *Logistic Regression*:

It is a supervised machine learning algorithm that accomplishes binary classification tasks by predicting the probability of an outcome, event, or observation [6].

- *Model Training and Evaluation*

The dataset was divided for training and evaluation using two cross-validation methods:
  - Leave-one-out cross-validation (LOOCV):
    Each instance in the dataset is used once as the test set while the remaining instances form the training set. This technique provides an unbiased estimate of model performance, particularly useful for small datasets.
  - K-fold cross-validation:
    The dataset is partitioned into k equally sized folds. Each fold serves as a test set while the remaining k-1 folds are used for training.

*D. Performance Evaluation and Validation:*

To evaluate the proposed method's performance, receiver operating characteristic (ROC) analysis was performed. The ROC curve represents the performance of a binary classifier as its discrimination threshold is varied [9]. It is a plot of the true positive rate (TPR) against the false positive rate (FPR) at various threshold settings. The true-positive rate is sensitivity, recall, or the probability of the correct diagnosis. The false-positive rate is the probability of a false diagnosis and can be calculated as 1 – specificity [9]. In this work, multiple receiver-operating-characteristic (ROC) curves were created for each predictive model, and the area under ROC curve was calculated. The classification accuracies were also calculated to evaluate performance of the proposed methods.

3. **RESULTS**

In this section the performance of the proposed methods is compared. The data were divided into training and validation sets using data partitioning and adjustment explained in the Methods section.

Data were collected from 30 healthy subjects, aged 19-37, recruited via advertisement at the University of Canterbury, under ethical approval and full informed consent. The trial included an even split of sex (15 male and 15 female), smokers/vapers (2 male and 6 female) and asthmatics (1 male and 1 female). Subject data were de-identified using subject numbers unassociated with signed consent forms. Demographic data were collected for each subject number (sex, height, weight, age, smoking history, history of asthma, and whether any of the trials made it feel harder or easier to breathe). **Figure 1** shows the corresponding signals for the same subject, including pressure, flow, tidal volume, and chest/abdominal circumference. **Figure 2** also shows V_tidal (tidal volume) waveforms for the three breathing types of a typical subject.

Breathing rate is estimated from the variables using an FFT algorithm and is used as an additional input variable for training machine learning algorithms. Regardless of whether pressure, flow, or tidal volume signals are used, the calculated breathing rate remains stable, with only minor variations, and the choice of signal does not significantly affect the accuracy of breathing rate estimation.

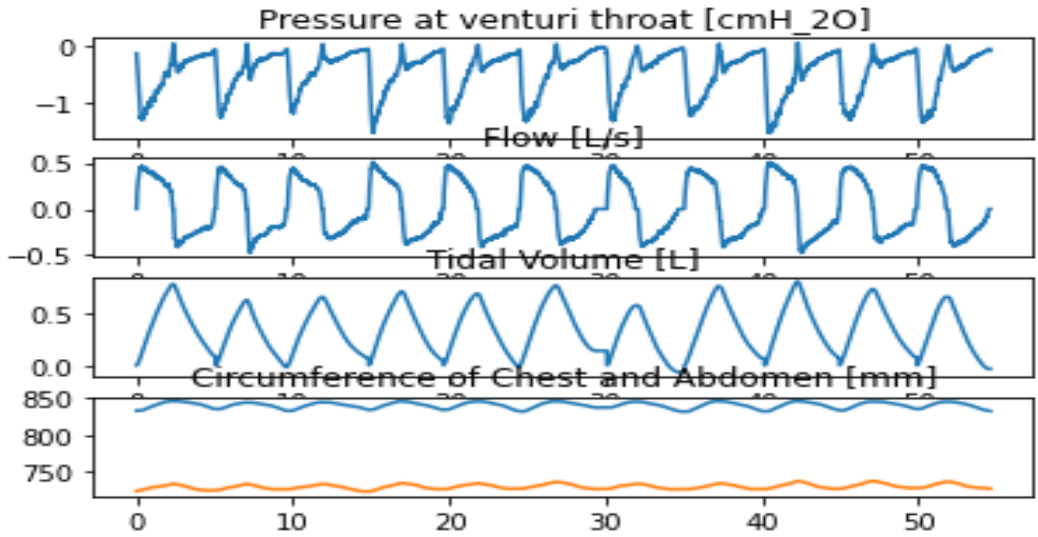

*Figure 1: Signals for the same subject as in Table 1.*

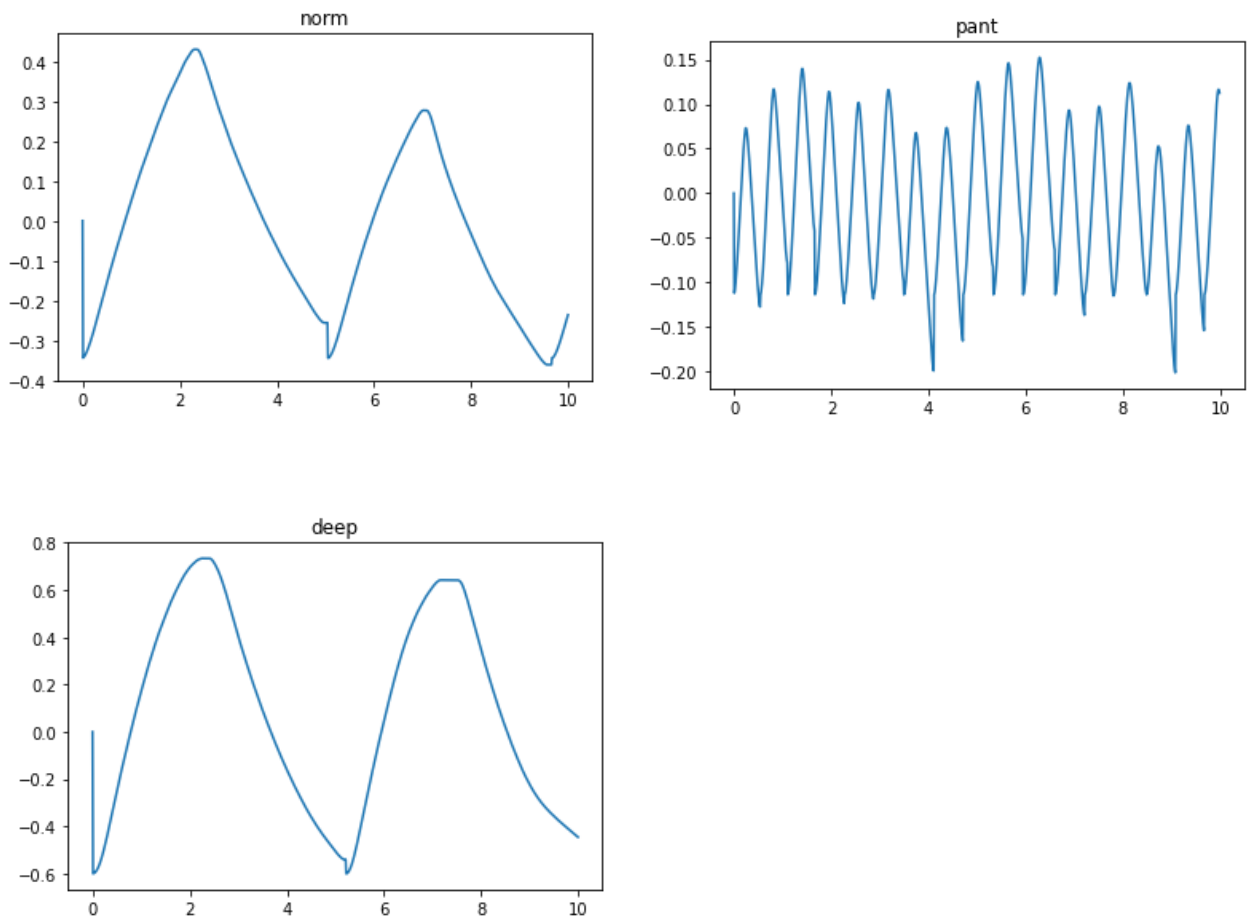

*Figure 2: Example V_tidal (Tidal Volume) waveforms for the three breathing types.*

Machine learning models were trained with this data using both LOOCV and K-fold cross validation for data partitioning and validation to predict breathing type [normal, panting, deep]. Prediction of breathing type was performed both with and without the breathing rate variable.

The processing steps implemented in the pipeline include:

- Data inspection/visualization: data are examined for quality assessment visually.
- Data pre-processing: remove NaN.
- Estimation of the breathing rate: used as an additional feature in AI training and for further clinical validation.
- Training various AI models: leave-one-out cross-validation and k-fold cross-validation.

The processed data includes time [s]; pressure [cmH2O]; flow [L/s]; tidal volume [L]; inspiratory start point indices; chest circumference [mm]; and abdominal circumference [mm]. Pressure, flow and volume can be used as inputs to pulmonary mechanics models to assess their fit and accuracy in identifying patient-specific parameters (e.g. lung elastance and airway resistance). Additionally, auxiliary dynamic circumferential monitoring allows for the secondary validation of these models as this data provides two separate methods of estimating changes in lung volume and muscular recruitment. The trial cohort includes an even split of males and females to assess sex-based differences in pulmonary mechanics. Asthmatics and smokers were also included in the trial for preliminary assessment of any pathophysiological implications for pulmonary mechanics and model-based assessment of these subjects. Also available is information on breathing rate type (normal, panting, or deep). We add a new feature, breathing rate (BR), estimated from existing measurements.

**Tables 1 and 2** demonstrate the top predictive performance for breathing patterns (normal, panting, deep) using leave-one-out cross-validation (LOOCV) and K-fold cross-validation respectively. In this work, both 5-fold and 10-fold cross-validation were performed to assess model robustness. According to these tables, when using LOOCV for data partitioning and validation, the accuracy for predicting breathing patterns is 88% when the estimated breathing rate is included as one of the input features, compared to 85% when it is not included. In contrast, using K-fold cross-validation, the best model performance is achieved with both the SVM and random forest models. The accuracy of breathing pattern prediction is around 75% with the random forest classifier showing optimal results at k=5 folds; it is 76% with SVM at k=10 folds. This suggests that regardless of the data partitioning and validation method—whether

LOOCV or K-fold cross-validation—random forest consistently outperforms other machine learning models for this clinical task.

*Table 1: Predictive performance of breathing pattern (normal, panting, deep) using leave-one-out cross-validation.*

| Performance | Breathing Pattern | |
|---|---|---|
| Machine Learning Model Type | Random Forest Classifier (without breathing rate feature) | Random Forest Classifier (with breathing rate feature) |
| Accuracy Mean | 85% | 88% |

*Table 2: K-fold cross-validation using different models for breathing type prediction. The breathing rate variable was not used in these models.*

| Performance | Random Forest Classifier (5-folds) | Logistic Regression (5-folds) | SVM with RBF kernel (10-folds) |
|---|---|---|---|
| Accuracy Mean | 75% | 42% | 76% |

## 4. DISCUSSION

Machine learning models, particularly the random forest classifier, can accurately predict breathing types based on non-invasive respiratory measurements obtained from CPAP therapy data. The high accuracy achieved using leave-one-out cross-validation underscores the model's effectiveness even with a limited sample size. The potential of these models aligns well with current guidelines and strategies for chronic obstructive lung disease (COPD), as outlined by the Global Initiative for Chronic Obstructive Lung Disease (GOLD) [10].

*Significance of the Breathing Rate Feature:*

The incorporation of breathing rate as an additional feature improved the model's performance. This finding aligns with physiological insights, as breathing rate is a critical indicator of respiratory health and can provide nuanced information beyond static measurements of pressure and flow. Previous research has similarly highlighted the importance of incorporating various physiological features to enhance predictive accuracy in respiratory monitoring [11].

*Model Performance Comparison:*

The random forest classifier outperformed other models, including logistic regression and SVM with an RBF kernel. The ensemble nature of random forest, which aggregates multiple decision trees, likely contributed to its superior ability to capture complex patterns in the data. This is consistent with findings from studies comparing machine learning models for respiratory health, where ensemble methods often demonstrated higher accuracy [12]. In contrast, logistic regression's linear approach was insufficient for the non-linear relationships inherent in respiratory data, as evidenced by its low accuracy.

*Limitations:*

Despite promising results, the work has several limitations:

- Sample Size: The dataset comprised only 30 participants, limiting the generalizability of the findings. A larger and more diverse population is necessary to validate the models robustly.
- Data Diversity: All data were from a single site (University of Canterbury), publicly available on PhysioNet [1], which may introduce site-specific biases. Multi-center data collection would enhance the model's applicability across different populations.

## 5. CONCLUSION

This work demonstrated method development and evaluation using machine learning-based algorithms for at-home respiratory disease monitoring and assessment in patients undergoing CPAP therapy. The random forest classifier, particularly when incorporating breathing rate as a feature, demonstrated high accuracy in predicting breathing types. These findings support the feasibility of transitioning respiratory assessments from specialized clinical environments to home settings, potentially enhancing accessibility and patient empowerment.

However, to fully realize this potential, further validation with larger, more diverse populations and the exploration of additional machine learning techniques is essential. Future work will focus on expanding the dataset, integrating additional health indicators, and conducting comprehensive clinical validation to establish the robustness and reliability of the proposed AI-driven respiratory monitoring system.